\documentclass[11pt,a4paper]{article}
\usepackage{times,latexsym}
\usepackage{url}
\usepackage[T1]{fontenc}

\newcommand{\method}{mFTI\ }
\newcommand{\methodend}{mFTI}
\usepackage{graphicx}
\usepackage{subcaption}
\usepackage{multirow}
\usepackage{amsmath}
\usepackage{inconsolata}
\usepackage{tabularray}
\usepackage{booktabs}
\usepackage{graphicx}
\usepackage{float}
\DeclareMathOperator*{\argmax}{argmax}
\newcommand{\green}[1]{\colorbox{green!30}{#1}}

\usepackage{CJK}
\usepackage{xcolor}
\usepackage{ulem}
\usepackage[utf8]{inputenc}

\newcommand\blfootnote[1]{%
\begingroup
\renewcommand\thefootnote{}\footnote{#1}%
\addtocounter{footnote}{-1}%
\endgroup
}
\usepackage[acceptedWithA]{tacl2021v1}

\usepackage{xspace,mfirstuc,tabulary}

\newif\iftaclinstructions
\taclinstructionsfalse 
\iftaclinstructions
\renewcommand{\confidential}{}
\renewcommand{\anonsubtext}{(No author info supplied here, for consistency with
TACL-submission anonymization requirements)}
\newcommand{\instr}
\fi

\iftaclpubformat 

\else

\fi

\title{Eliciting the Translation Ability of Large Language Models via Multilingual Finetuning with Translation Instructions}

\author{
  Jiahuan Li$^{1*}$, Hao Zhou$^{1*}$, Shujian Huang$^{1\dagger}$, Shanbo Cheng$^2$ \and
  Jiajun Chen$^1$
  \\
  \ \\
  $^1$ National Key Laboratory for Novel Software Technology, Nanjing University, China
\\
  \texttt{\{lijh,zhouh\}@smail.nju.edu.cn, \{huangsj,chenjj\}@nju.edu.cn}
  \\
\\
  $^2$ Bytedance Research
\\
  \texttt{chengshanbo@bytedance.com}
}

\date{}

\begin{document}
\maketitle
\blfootnote{* Equal contribution.}
\blfootnote{$\dagger$ Corresponding author.}
\begin{abstract}
Large-scale Pretrained Language Models~(LLMs), such as ChatGPT and GPT4, have shown strong abilities in multilingual translation, without being explicitly trained on parallel corpora. It is intriguing how the LLMs obtain their ability to carry out translation instructions for different languages. In this paper, we present a detailed analysis by finetuning a multilingual pretrained language model, XGLM-7.5B, to perform multilingual translation following given instructions. Firstly, we show that multilingual LLMs have stronger translation abilities than previously demonstrated. For a certain language, the translation performance depends on its similarity to English and the amount of data used in the pretraining phase. Secondly, we find that LLMs' ability to carry out translation instructions relies on the understanding of translation instructions and the alignment among different languages. With multilingual finetuning with translation instructions, LLMs could learn to perform the translation task well even for those language pairs unseen during the instruction tuning phase.
\end{abstract}

\section{Introduction}

The emergence of Large Pretrained Language Models (LLMs) \citep{gpt3,chatgpt} has revolutionized the research of machine translation \citep{hendy2023good,garcia2023unreasonable}. These models have demonstrated remarkable multilingual translation capabilities, without requiring explicit training on parallel corpora. For instance, XGLM, a medium-sized multilingual language model, outperforms supervised models using only several examples as demonstrations \citep{lin-etal-2022-shot}; the cutting-edge LLM GPT4 has been shown to perform comparably to commercial translation systems on multiple language pairs \citep{jiao2023chatgpt}.

Most existing research on LLMs for machine translation focuses on in-context learning (ICL), i.e. taking several parallel sentences as the demonstration to guide LLMs to perform translation \citep{vilar-etal-2023-prompting,agrawal-etal-2023-context,hendy2023good,zhu2023multilingual}. However, these methods rely heavily on the in-context learning ability of LLMs. For smaller models, e.g. models with only 1B or 7B parameters, the relatively weak ICL ability may result in an underestimation of their potential translation ability.

Instead of relying on the ICL abilities, we propose to investigate the ability of LLMs by directly training them to follow translation instructions.
Inspired by the recent success of instruction tuning~\citep{wei2022finetuned,chung2022scaling}, we organize multilingual translation tasks as different instances of the translation instruction, with each instance corresponding to a specific language pair. By training the LLMs to follow these instructions, i.e.  with \textbf{m}ultilingual \textbf{F}inetuning with \textbf{T}ranslation \textbf{I}nstructions (\methodend),
it is possible to better elicit translation ability inside LLMs.

Our results show that by training on a mixed dataset of 1,000 sentences per language pair, \method outperforms the 8-shot in-context learning by near 3 BLEU on average, showing a greater potential of LLMs' translation ability than previously demonstrated~\citep{lin-etal-2022-shot}. In addition, we also discuss how \method improves the LLMs and which factors influence the performance.

To better understand why LLMs could follow these instructions, we design a \method setting where only a subset of the translation instructions, i.e. language pairs, are used for training. Thus LLMs need to generalize their instruction following abilities for those language pairs unseen during \methodend. Surprisingly, \method elicits the translation ability not only for trained language pairs but also for those unseen during instruction training. With further experiments and analyses, we find that LLMs could learn the translation behavior in general by being trained to translate even irrelevant language pairs. It is also interesting that with \methodend, LLMs learn to directly align languages through the use of pivot languages, which enhances the instruction-following ability for unseen language pairs.

\section{Multilingual Finetuning with Translation Instructions}
\label{sec:background}

\subsection{Overall Framework}

Given a corpus of multilingual parallel sentences and their languages $\mathcal{M} = \{({l_s}^i, {l_t}^i, \mathbf{x}^i, \mathbf{y}^i) \}$, where ${l_s}^i$ and ${l_t}^i$ are names of the source and target language of $i$-th parallel sentence $(\mathbf{x}^i, \mathbf{y}^i)$, respectively, \method leverages an instruction template $\mathcal{T}$ to organize the corpus $\mathcal{M}$ into a language modeling dataset $\mathcal{D}$. Each sentence $d^i$ in $\mathcal{D}$ is an instantiation of the translation instruction with a specific sentence pair: ${d}^i = \mathcal{T}({l_s}^i, {l_t}^i, \mathbf{x}^i, \mathbf{y}^i)$. The parameter of LLMs are then optimized using a standard next-token-prediction objective on $\mathcal{D}$:

\begin{equation}
    \underset{\theta}{\argmax}  \sum_{i=1}^{|\mathcal{D}|} \sum_{j=1}^{|{d}^i|} \texttt{log}\ p_{\theta}( {d}_j^i | {d}_{<j}^i),
\end{equation}

\noindent where $\theta$ are parameters of LLMs. The instruction template we adopt is 
\begin{equation*}
    \text{Translation: [}{l_s}\text{]: }\mathbf{x}\text{ [}{l_t}\text{]: }\mathbf{y}
\end{equation*}

\noindent where the prefix ``Translation:'' is used to indicate the translation task; the pattern ``[$\cdot$]:'' is used to identify the name of the specific language. 

\subsection{Experiment Setup}

\paragraph{Backbone Language Model}
We consider XGLM-7.5B \citep{lin-etal-2022-shot} as our backbone language models.  XGLM-7.5B is a massive multilingual auto-regressive language model, which is trained on a massive corpus of 500 billion tokens comprising 30 diverse languages. Low-resource languages have been up-sampled during training, making it an ideal backbone model for multilingual translation research.

\paragraph{Languages}
Following \citet{lin-etal-2022-shot}, our evaluation involves 13 languages that are covered in the pretraining corpus of XGLM, i.e. English~(En), German~(De), French~(Fr), Catalan~(Ca), Finnish~(Fi), Russian~(Ru), Bulgarian~(Bg), Chinese~(Zh), Korean~(Ko), Arabic~(Ar), Swahili~(Sw), Hindi~(Hi) and Tamil~(Ta). Among these languages, En, De, Fr, Ru, and Zh are high-resource languages (with ratios in the XGLM pretraining data greater 4\%); Ko, Fi, Ar, Bg are medium-resource languages (with ratios between 0.5\%-4\%); Ca, Hi, Ta, Sw are low-resource languages (with ratios under 0.5\%)

\paragraph{Evaluation Datasets}
Following previous works~\citep{lin-etal-2022-shot}, we evaluate translation models on the FLORES-101 dataset~\citep{goyal-etal-2022-flores}, which provides manual translations of 1012 sentences in 101 languages. 

\paragraph{Finetuning Datasets}
Our finetuning dataset primarily comes from WikiMatrix \citep{wikimatrix}. WikiMatrix provides a parallel corpus for 1620 different language pairs, including many non-English language pairs, which enables a systematic investigation for the translation of languages other than English. We also leverage the MultiCCAligned \citep{elkishky_ccaligned_2020} corpus for language pairs that are not contained in WikiMatrix, including Hi-Sw, Ko-Sw, Ta-Sw, Sw-Hi, Sw-Ko, Sw-Ta.

\paragraph{Optimization Details}
We finetune all models using the Adam \citep{kingma2017adam} optimizer with the learning rate fixed as $5e-6$. We use a fixed batch size of 80 sentences and finetune models for 1 epoch or 2000 steps (depending on the size of the training corpus) for all experiments.

\section{Understanding the Potential Translation Ability of LLMs}
\label{sec:finetuning}
In this section, we first assess the overall translation performance of \method by comparing it to few-shot in-context learning\footnote{We randomly select 8 examples from the FLORES-101 dev split as the demonstration for ICL. Random selection strategy is shown to be good enough in many previous works~\citep{vilar-etal-2023-prompting,zhu2023multilingual}. The template we use for ICL is 
 <src\_text> = <tgt\_text>, which shows good performance according to~\citet{zhu2023multilingual}.}. We then present a detailed analysis of how the corpus for \method influences the translation quality.

\begin{figure*}[t]
  \centering
  \includegraphics[width=1.0\linewidth]{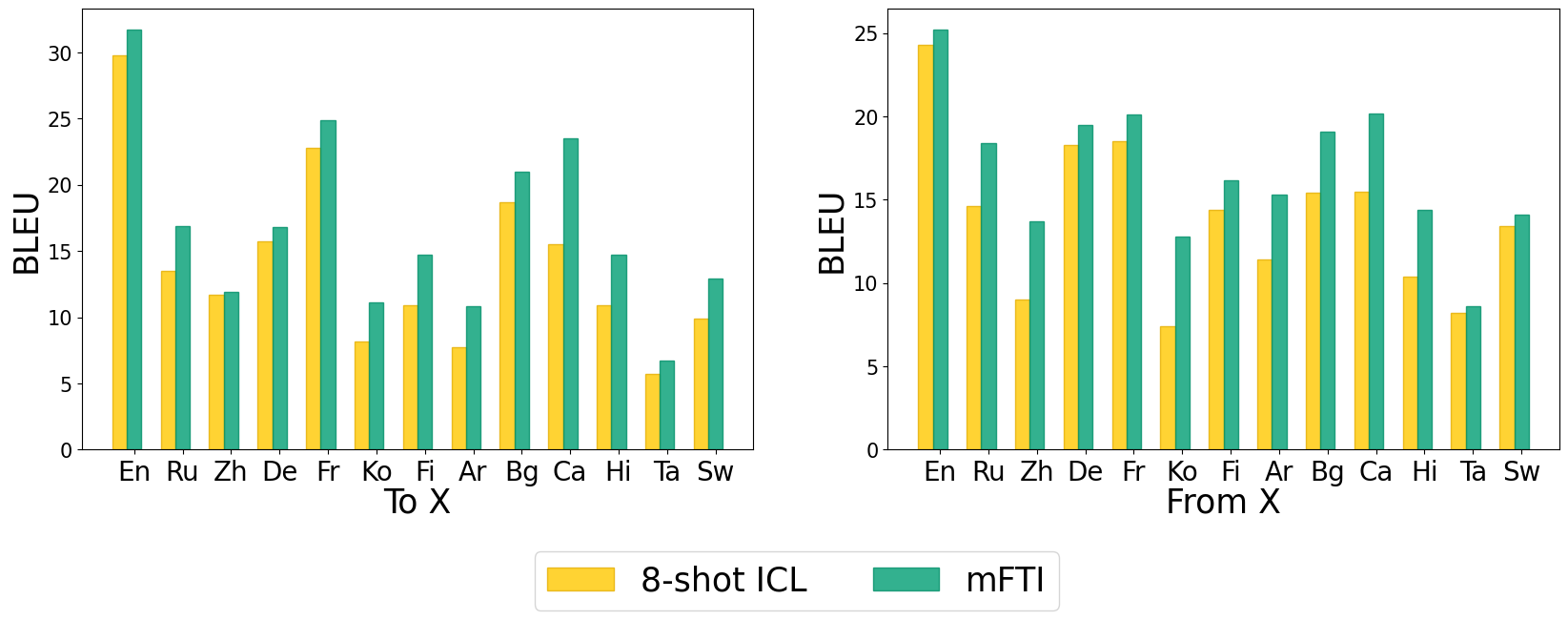}
  \caption{Translation performance of 8-shot ICL and \method using 1000 sentences per language pair. Languages are ordered by the data amount in the pretraining corpus.}
  \label{fig:main_results}
\end{figure*}

\subsection{Translation Ability of LLMs}

We finetune XGLM on 156 language pairs spanning all 13 languages. Since our goal is to elicit the translation ability of LLMs using a small number of examples, we limit the number of parallel sentences to 1000 per language.

\paragraph{\method Better Elicits Translation Ability than Few-shot ICL.} Figure \ref{fig:main_results} shows the average BLEU for translation to and from language X, respectively. Full results on each langauge direction can be found in Appendix~\ref{appendix: full results}.
It is clear that \method leads to better translation performances than 8-shot ICL for all language pairs (3 BLEU on average). For some languages, the gap is up to 8 BLEU (e.g. translating into Catalan). This demonstrates the effectiveness of \method in eliciting LLM's translation ability. It also shows that LLMs have a greater potential for multilingual translation than we saw with ICL~\citep{lin-etal-2022-shot}. 

Even for translating to and from English, \method still outperforms 8-shot ICL, but with a much smaller gap. This indicates that LLMs with ICL are better at performing tasks that involve English rather than other languages, but they still have the potential to perform even better. 

\begin{table}[t]
\centering
\begin{tabular}{@{}rcc@{}}
\toprule
\multicolumn{1}{l}{}                           & \textbf{To X} & \textbf{From X} \\ \midrule
\multicolumn{1}{l}{\textbf{Data Amount in Pretraining}} & 0.39          & 0.36            \\ \midrule
\multicolumn{3}{l}{\textbf{Similarity To English}}                                        \\
\textit{Geography}                                      & 0.93          & 0.87            \\
\textit{Syntax}                                         & 0.85          & 0.80            \\
\textit{Phylogeny}                                      & 0.71          & 0.75            \\
\textit{Phonology}                                      & 0.50          & 0.49            \\
\textit{Inventory}                                      & 0.51          & 0.41            \\ \bottomrule
\end{tabular}
\caption{Spearman correlation between average translation performance (in BLEU) and possible influence factors (data amount in pretraining, similarity to English). The performance of translating to and from language X is calculated separately.}
\label{tab:factor}
\end{table}

\paragraph{XGLM is still an English-centric Model.}
The translation performance for each language varies greatly. Considering that the number of sentences used in \method is the same for each language, one may suspect that the translation performance of each language largely depends on the amount of its pretraining data. For this reason, the languages in Figure \ref{fig:main_results} are listed in descending order of their data amount in the XGLM pretraining. However, there are clear fluctuations. For example, Russian and Chinese are the two languages with the largest portion of pretraining data other than English, but their translation performance is much worse than some other languages such as French. 

We calculate the Spearman correlation between the translation performance and possible influence factors, namely data amount in pretraining and similarity to English. For data amount, we use the size of the pretraining corpus reported in \citet{lin-etal-2022-shot}. For similarity to English, we adopt the lang2vec\footnote{\url{https://github.com/antonisa/lang2vec}}, which is a toolkit for querying the URIEL typological database, to get each language's feature vector of different perspectives including  \textit{geography}, \textit{syntax}, \textit{phylogeny}, \textit{phonology} and \textit{inventory}\footnote{We refer readers to~\citet{littell-etal-2017-uriel} for details on how the feature vector is obtained.}. 

As shown in Table~\ref{tab:factor}, the translation performance indeed has a positive correlation with data amount in pretraining (0.39/0.36). However, the similarity between a specific language and English plays a more important role in determining the final translation performance. All considered features demonstrate a higher correlation coefficient than the data amount in pretraining. This indicates that XGLM is still a predominantly English-centric model. Based on these observations, we suggest taking the relation between different languages into consideration when collecting and sampling data for pretraining multilingual LLMs.

\begin{figure}[t]
\centering
\includegraphics[width=0.95\linewidth]{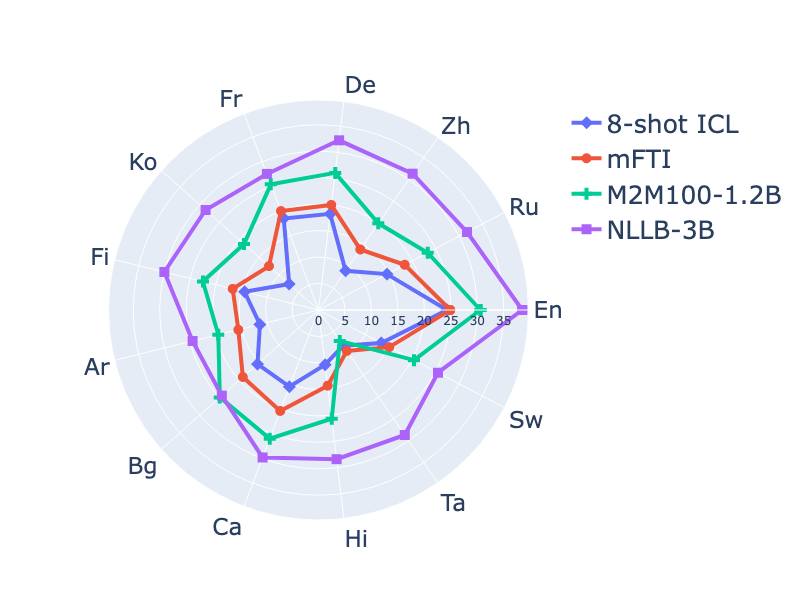}
\caption{Comparison of \method with conventional supervised machine translation models. Performances are evaluated in BLEU. }
\label{fig:supervised}
\end{figure}

\begin{figure}[t]
\centering
\includegraphics[width=0.95\linewidth]{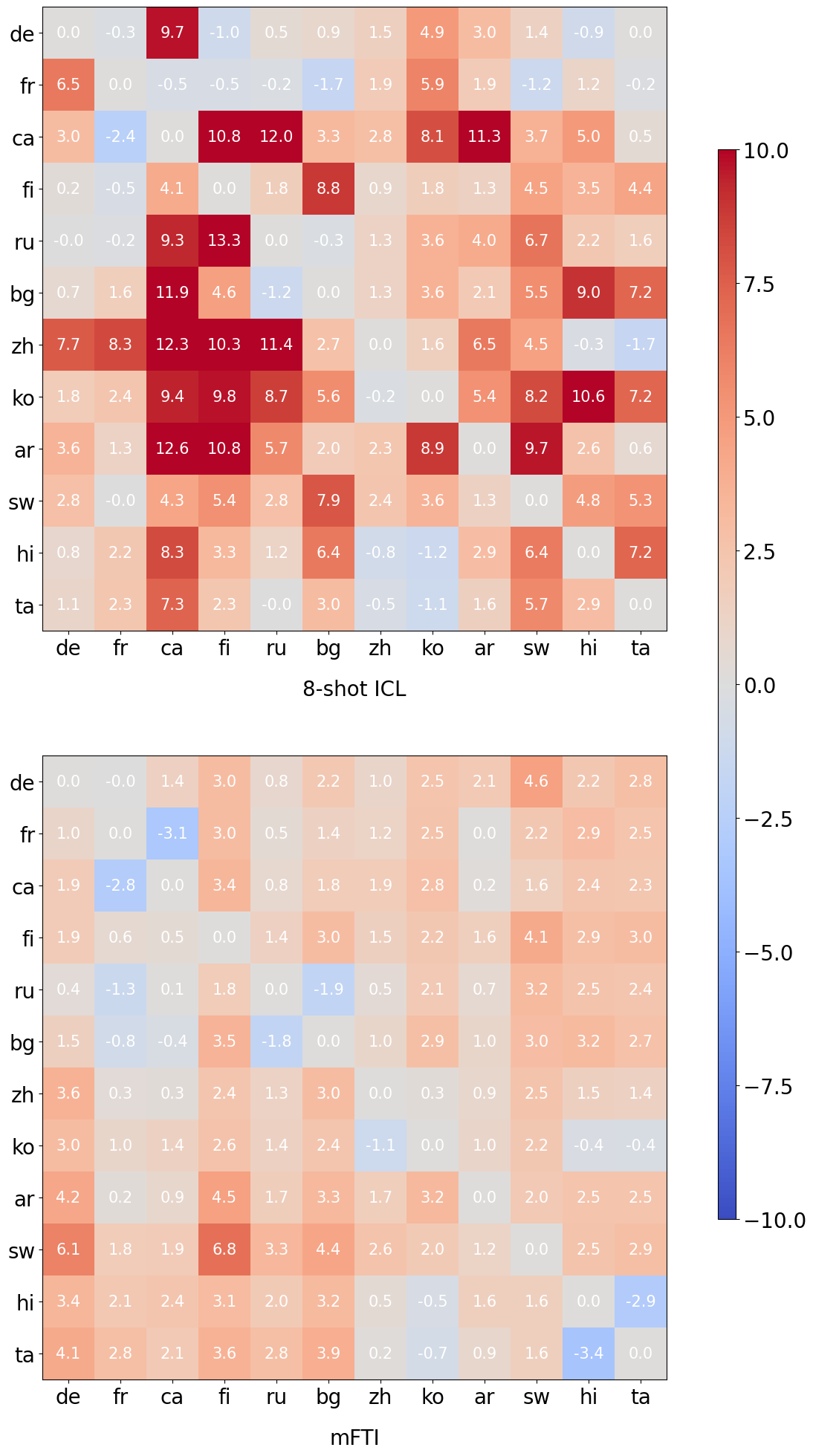}
\caption{Changes of BLEU score after pivoting through English for 8-shot ICL and \methodend.}
\label{fig:pivot}
\end{figure}

\paragraph{It is not trivial for LLM-based models to outperform conventional supervised MT models.} To better posit the performance of \method, we compare it with two conventional supervised MT models, i.e., M2M-1.2B~\citep{m2m100} and NLLB-3B~\citep{nllb} model, in Figure~\ref{fig:supervised}~\footnote{We also include the performance evaluated by COMET in Appendix~\ref{sec:supervised_comet}.}. We can see that despite that \method significantly improves over 8-shot ICL and sometimes achieves comparable performance to M2M-615M, it still lags behind the stronger NLLB-3B by a large margin, rendering the challenge to adopt a medium-sized LLM to outperform large-scale supervised MT models.

\subsection{\method Brings Consistent Improvements across Different Metrics, LLMs and Finetuning Strategies}
In order to understand the universal effectiveness of \methodend, we present experiments on more LLMs, i.e. BLOOM-7b1~\citep{scao2022bloom} and LLaMA~\citep{touvron2023llama}, and parameter-efficient finetuning strategy LoRA~\citep{hu2022lora}. We report the performance averaged on 156 translation directions evaluated by both sacreBLEU~\citep{post-2018-call} and COMET~\citep{rei-etal-2022-comet}~\footnote{We use the \textrm{wmt22-comet-da} version.} in Table~\ref{tab:other llms}~\footnote{Detailed hyperparameters are in Appendix \ref{appendix:hyperparameters}.}. 

Firstly, we can see that methods based on XGLM-7.5B significantly performs significantly better than BLOOM-7B and LLaMA-7B. This is because many low-resource languages are ill-represented in BLOOM and LLaMA. Secondly, \method consistently outperforms 8-shot ICL in terms of BLEU and COMET on all three studied LLMs, regardless of the finetuning strategy, which demonstrates the universal effectiveness in different scenarios. Contrary to previous findings~\citep{jiao2023parrot}, we did not find LoRA performs better than full finetuning. We hypothesize that learning translation on 156 pairs simultaneously is more challenging and requires more model capacity, making full finetuning a better choice than LoRA in this scenario.

\begin{table*}
\centering
\begin{tabular}{lcccccc}
\toprule
   & \multicolumn{2}{c}{\textbf{BLOOM-7B}} & \multicolumn{2}{c}{\textbf{LLaMA-7B}}  & \multicolumn{2}{c}{\textbf{XGLM-7.5B}}  \\ \midrule
   &  \textbf{BLEU}       & \textbf{COMET}    &\textbf{BLEU}       & \textbf{COMET}   &\textbf{BLEU}       & \textbf{COMET}   \\
8-shot ICL        & 8.4       & 60.9       & 9.0       & 61.0    & 13.9    & 73.4    \\
\method~(LoRA)  & 9.0      & 64.3       & 9.5       & 63.9    & 16.7    & 77.0  \\ 
\method~(Full Finetuning)       & \textbf{10.2}       & \textbf{65.4}       & \textbf{9.8}       & \textbf{66.0}    & \textbf{16.9}    & \textbf{77.7}  \\ \bottomrule
\end{tabular}
\caption{Averaged translation performance on all 156 language pairs of 8-shot ICL and \method using different LLMs and finetuning strategies. }
\label{tab:other llms}
\end{table*}

\subsection{\method Enhances Direct Language Alignment} 
A distinct difference between ICL and \method is that \method could learn from more parallel sentences and update the model if needed. It is interesting to see what changes after the update. Many previous works \citep{zhang2023prompting,jiao2023chatgpt} have shown that translating by pivoting through English significantly improves ICL's translation performance. To this end, we compare performance gains of pivot translation using ICL and \methodend, respectively. 

Figure~\ref{fig:pivot} presents the result. Each value in the grid is the BLEU difference before and after pivoting through English. We can first observe that pivoting through English indeed improves translation performance for ICL, up to 10 BLEU in some language pairs. However, after \methodend, the gap has been significantly reduced. Considering the fact the \method achieves an average 3 BLEU higher than ICL, the reduction of benefits from pivoting through English compared to direct translation may indicate a better direct alignment between languages.

\begin{table}[t]
\centering
\begin{tabular}{rcc}
\toprule
\multicolumn{1}{l}{}  & \textbf{BLEU}  \\ \midrule
Low quality      & 15.0         \\
High quality & 16.9       \\ \bottomrule
\end{tabular}
\caption{The translation performance of finetuned XGLM as the quality of finetuning corpus varies.}
\label{tab:low}
\end{table}

\subsection{Influencing Factors of \method}
\paragraph{Quality of Finetuning Corpus is Crucial.}

Recent work on instruction tuning demonstrates that the quality of instruction data is crucial for achieving good performances~\citep{zhou2023lima}. We observe a similar trend when performing \methodend. Specifically, we construct high and low-quality finetuning corpus by selecting parallel sentences according to their attached LASER\footnote{\url{https://github.com/facebookresearch/LASER}} similarity score from the full set of parallel sentences. According to the results in Table \ref{tab:low}, finetuning with high-quality parallel sentences can improve the BLEU score by around 2 points compared to finetuning with low-quality parallel sentences, emphasizing the importance of corpus quality, validating the importance of the quality of finetuning corpus.

\begin{figure}[t]
\centering
\includegraphics[width=0.85\linewidth]{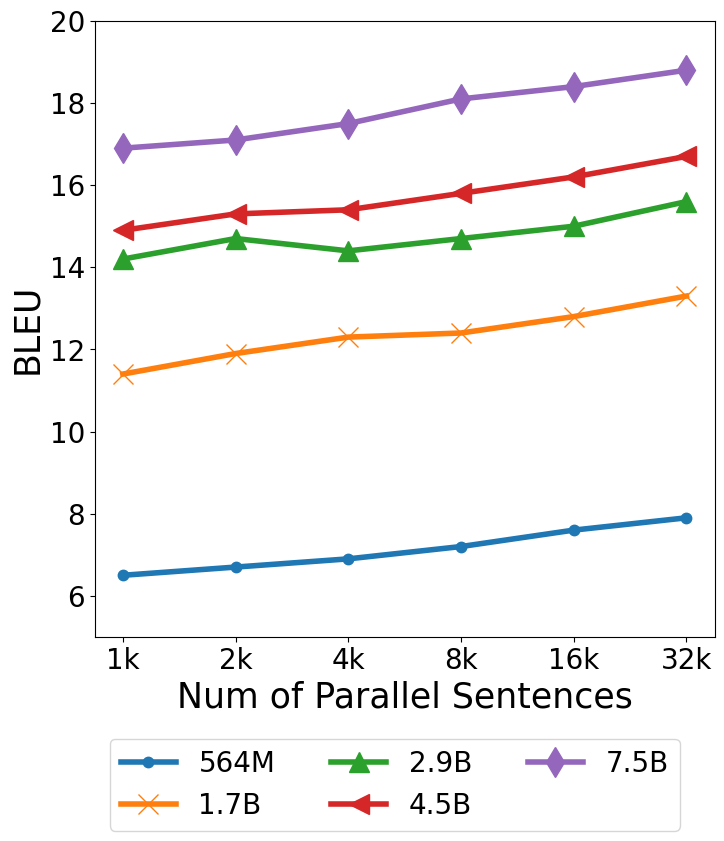}
\caption{The translation performance of finetuned XGLM as the number of model parameters and training examples scales.}
\label{fig:scaling}
\end{figure}

\paragraph{The Effectiveness of \method Scales with Model size and Training Examples.}
Figure \ref{fig:scaling} shows the translation performance when varying the number of training examples per language pair (1k, 2k, 4k, 8k, 16k, 32k) and the number of model parameters (564M, 1.7B, 2.9B, 4.5B, 7.5B). As we can see, it follows a standard log-linear scaling law in terms of both the number of training examples and model size, which is consistent with findings in the previous work \citep{kaplan2020scaling}.

\begin{table*}[t]
\footnotesize
\centering
\begin{tabular}{@{}cllllll@{}}
\toprule
                  & \multicolumn{3}{c}{\textbf{Seen Both Sides}}                                       & \multirow{2}{*}{\textbf{Unseen Src}} & \multirow{2}{*}{\textbf{Unseen Tgt}} & \multirow{2}{*}{\textbf{Unseen Both Sides}} \\ 
                  & \textbf{Same Direction} & \textbf{Reversed Direction} & \textbf{Unseen Direction} &                                      &                                      &                                       \\ \midrule
\textbf{8-shot ICL}   & 14.5                    & 14.5                        & 11.2                      & 13.5                                 & 13.6                                 & 14.6                                  \\
\textbf{\methodend-16}  & 15.7(+1.2)                   & 13.7(-0.8)                        & 12.6(+1.4)                      & 14.9(+1.4)                                 & 14.5(+0.9)                                 & 15.3(+0.7)                                  \\ \midrule
\textbf{\methodend-all} & 16.7                    & 16.8                        & 14.6                      & 17.6                                 & 17.0                                 & 18.4                                  \\ \bottomrule
\end{tabular}
\caption{Translation performances under  different data conditions. \methodend-16: XGLM multilingual finetuned with translation instructions on a mixture of 16 language pairs described in Section~\ref{sec:conditions}.}
\label{tab:partial}
\end{table*}

\section{Understanding the Ability of Carrying Out Translation Instructions}
In this section, we present a comprehensive analysis on how \method improves the model's ability to carry out translation instructions.

We begin by presenting an overarching experiment where we intentionally withhold certain language pairs during the \method process, which allows us to study models' ability to carry out translation instructions under different conditions.

Furthermore, we delve deeper into our analysis by exploring how \method enhances LLMs' ability to 
carry out translation instructions from following perspectives: better understanding of translation instructions~(Section~\ref{sec:partial_scaling} and Section~\ref{sec:translation_behavior}) and better alignment between languages to execute translation instructions~(Section~\ref{sec:language_alignment}).

\subsection{Manipulating Conditions}
\label{sec:conditions}
In Section \ref{sec:finetuning}, we have presented results in a fully supervised setting, where all testing language pairs are seen during instruction tuning.  To provide further insights into LLMs' generalization ability across language pairs, we simulate a more realistic scenario where there may be a lack of source and/or target language sentences during the instruction tuning process.

More specifically, from the 13 selected languages, we hold out 6 languages as unseen languages.
We further partition the rest 7 languages into three groups: Only-Source (languages only appear on the source side), Only-Target (languages only appear on the target side) and Source-Target (languages appear on both the source and target side). We then form language pairs from these partitions following the requirement of partitions. This allows us to assess \methodend's performance under the following conditions:

\begin{itemize}
    \item \textbf{Seen Both Sides} Both the source side and target side language appear in the finetuning corpus. This can be further divided to:
        \begin{itemize}
            \item \textbf{Same Direction}. The same translation direction is trained during \methodend.
            \item \textbf{Reversed Direction}. The same translation direction does not appear when training, but the reversed direction does.
            \item \textbf{Unseen Direction}. The translation pair (neither the same nor the reverse) does not appear when training.
        \end{itemize}
    \item \textbf{Unseen Src}. Only the target language sentences appear when training.
    \item \textbf{Unseen Tgt}. Only the source language sentences appear when training.
    \item \textbf{Unseen Both Sides}. Neither source language nor target language sentences appear in the finetuning corpus.
\end{itemize}

\subsection{\method Learns to Follow Translation Instruction across Conditions}
\label{sec:partial}

We finetune XGLM on the corpus described in the previous section. Since there are 16 language directions in the training corpus, we denote the finetuned model as \methodend-16. The model finetuned on all language pairs is denoted as \methodend-all. Table~\ref{tab:partial} shows the results.  

\paragraph{\methodend-16 Brings Improvements on Most Settings, Yet Much Less Than \methodend-all.}Firstly we can see that \methodend-16 brings improvements on most settings except Reversed Direction, demonstrating the effectiveness of \methodend-16. However, the improvements are less when compared \methodend-all, even for the Same Direction partition. This can be attributed to fewer language pairs when finetuning, which we will discuss in Section \ref{sec:partial_scaling}.

\paragraph{Language Position Shift Between Training and Testing Has Negative Effects on Translation Performance.}
The translation performance of \methodend-16 on Reversed Direction degrades by 0.8 BLEU compared to 8-shot ICL. By inspecting the translation results, we find that \methodend-16 suffers from severe off-target problems, i.e. generating translations in wrong target languages. We hypothesize that this could be attributed to the shift in the relative positions of the source and target languages during training.

\paragraph{Seeing Target Languages When Finetuning is Better Than Source Languages.}
When there are unseen languages in the language direction, the improvement on Unseen Src is much larger compared to Unseen Tgt, indicating the understanding of the specified target language may be more important than the source language.

\paragraph{Unseen Both Sides Also Benefit From \method Training.} The most surprising phenomenon is that language pairs from Unseen Both Sides partition also benefit from \methodend, with an improvement of 0.7 BLEU compared to 8-shot ICL. Since \methodend-16 does not see any sentences of the source and target languages, the improvements indicate a better understanding of the translation instruction, which we will discuss in Section \ref{sec:translation_behavior}.

\begin{figure}[t]
\centering
\includegraphics[width=1.0\linewidth]{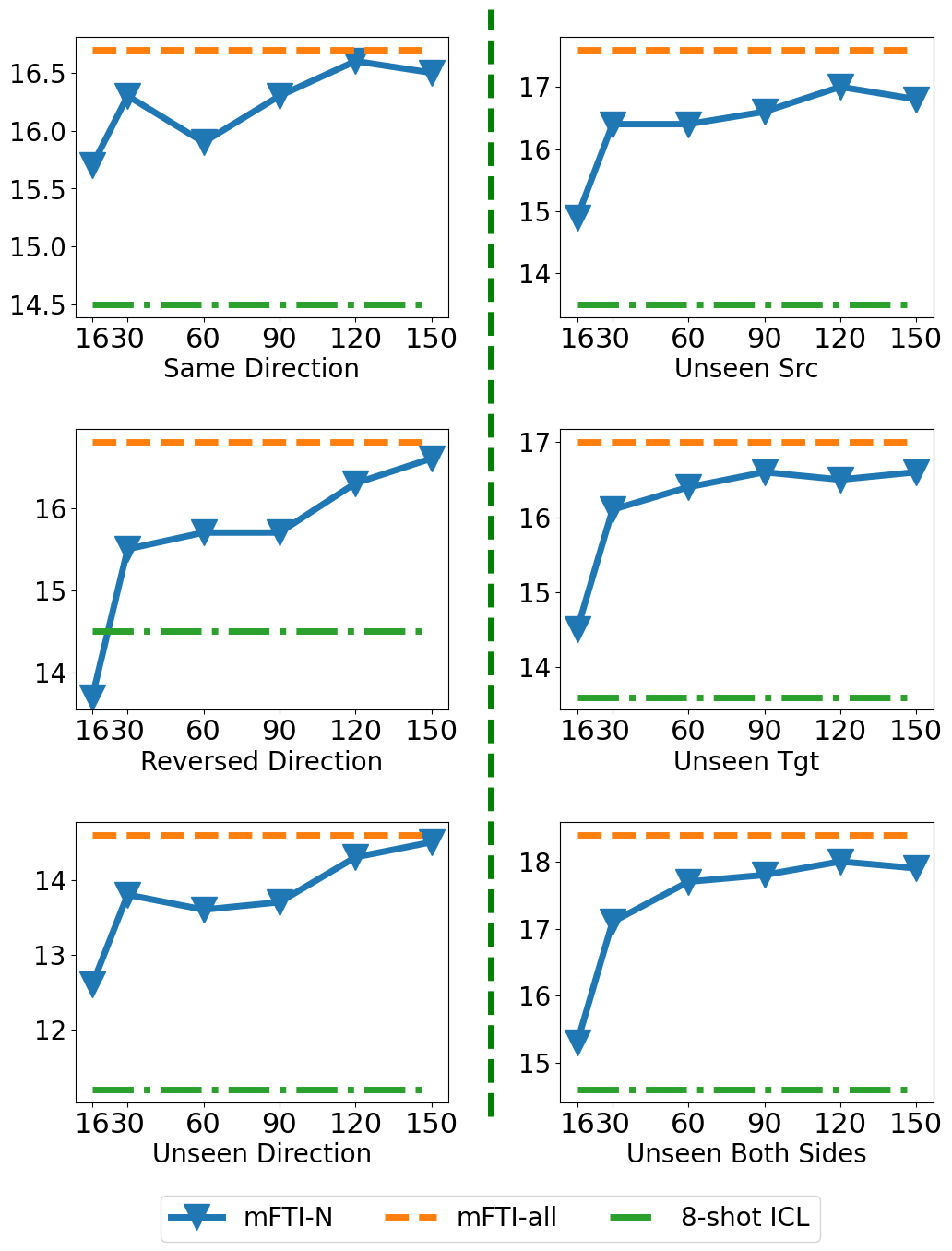}
\caption{Translation performance on different partitions as the number of language pairs grows. \textit{Left}: partitions where sentences of both source and target language are seen when training. \textit{Right}: partitions where source and/or target language sentences are unseen when training. }
\label{fig:partial_scaling}
\end{figure}

\begin{figure*}[t]
\centering
\includegraphics[width=1.0\linewidth]{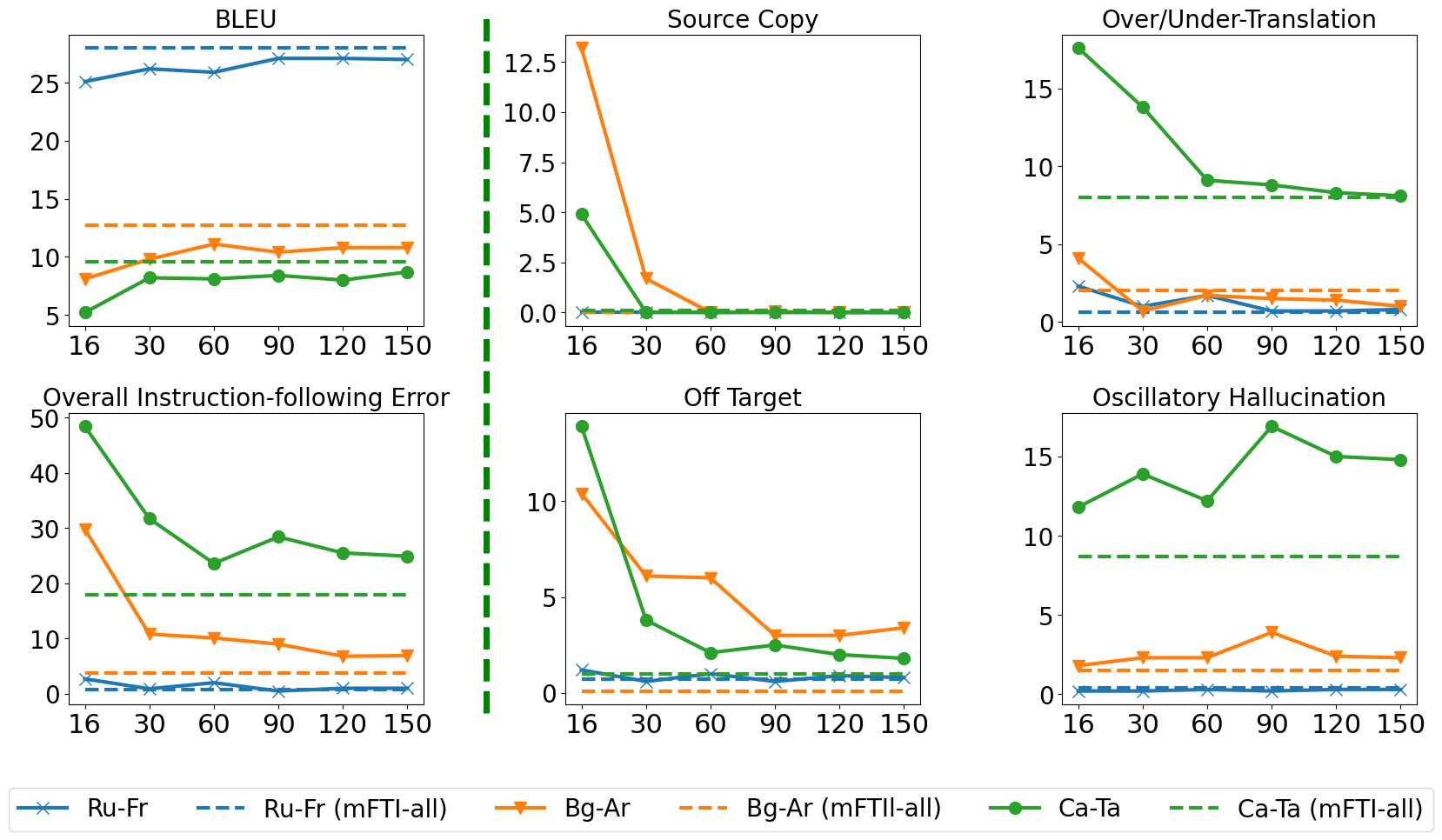}
\caption{Trends of translation and instruction-following performance on 3 Unseen-both language pairs when scaling up the number of language pairs during mFTI. The left 2 figures show the BLEU score and overall instruction-following error ratios, respectively. The rest 4 figures show the ratios of 4 specific error types, respectively, i.e. source copy, off-target, over/under translation, and oscillatory hallucination. The X-axis denotes the number of training language pairs. The Y-axis denotes the percentage of translations with specific error types.}
\label{fig:add_language_pairs}
\end{figure*}

\subsection{Instruction Tuning with More Language Pairs Leads to Better Translation Performance}
\label{sec:partial_scaling}
Previous instruction-tuning works show that scaling the number of tasks significantly benefits the unseen tasks~\citep{chung2022scaling}. Observing the performance gap of Same Direction between \methodend-16 and \methodend-all, we gradually add more language pairs to \methodend-16, and plot the translation performance on each partition in Figure~\ref{fig:partial_scaling}. In order to isolate possible effects of additional monolingual sentences, we only add language pairs that exclude the studied 13 languages~\footnote{Detailed language pairs are in Appendix \ref{appendix:Additional language}.}.

It can be seen that as the number of language pairs grows, the translation performances of all partitions generally increase, validating the importance of more language pairs. Notably, the performance of the Reversed Direction partition is significantly boosted, outperforming 8-shot ICL by a large margin when increasing the number of language pairs from 16 to 30.

Surprisingly, the performance of the Unseen Both Sides partition improves the most. Since no data of language pairs in Unseen Both Sides are added, this indicates the ability of instruction-following on these language pairs has been significantly enhanced, which we will discuss in the next section.

\subsection{\method Generalizes the Understanding of Translation Instruction to Unseen Directions}
\label{sec:translation_behavior}

In this section, we aim to understand how \method facilitates the understanding of instructions from a more fine-grained view, i.e. specific \textit{language directions} and \textit{instruction-following errors}. 

For the language directions, we select Ru$\to$Fr~(high-resource), Bg$\to$Ar~(medium-resource), Ca$\to$Ta~(low-resource) from the Unseen-Both Sides partition to study \methodend's effectiveness under different resource settings. 

For instruction errors, we identify the following four major problems in translations:

\begin{itemize}

    \item \textit{Source Copy}~(\textbf{SC}): This error occurs when the model simply copies the source sentence as the translation without making any meaningful changes. We identify this error by calculating the sentence-level BLEU score between the translations and the source sentences. If the BLEU score is above 80, it indicates that the translation is nearly identical to the source.

    \item \textit{Off-target translation}~(\textbf{OT}): In this case, the model fails to produce sentences in the target language. We detect this error by using a language identification tool, such as \textit{fasttext}, to determine the language of the generated translations.

    \item \textit{Over/under translation}~(\textbf{OU}): This error refers to situations where the model produces translations that are significantly longer or shorter than references. We consider translations with a length ratio above 2 or below 0.5 as over- or under-translations, respectively.

    \item \textit{Oscillatory hallucination}~(\textbf{OH}): This error occurs when the model gets stuck in a specific translation state and generates repeated n-grams until reaching the maximum length. We define translations with n-grams that consecutively repeat at least three times as oscillatory hallucinations.
\end{itemize}

\begin{table*}[t]
\centering
\begin{tabular}{rccccccccc}
\toprule
                       & \multicolumn{3}{c}{\textbf{Ru$\to$Fr}} & \multicolumn{3}{c}{\textbf{Bg$\to$Ar}} & \multicolumn{3}{c}{\textbf{Ca$\to$Ta}} \\ \midrule
                       & \textbf{OT$\Downarrow$}        & \textbf{OH$\Downarrow$}        & \textbf{BLEU$\Uparrow$}       & \textbf{OT$\Downarrow$}     & \textbf{OH$\Downarrow$}     & \textbf{BLEU$\Uparrow$}    & \textbf{OT$\Downarrow$}      & \textbf{OH$\Downarrow$}     & \textbf{BLEU$\Uparrow$}   \\
mFTI-16        & 1.2       & 0.2       & 25.1       & 10.4    & 1.8    & 8.1     & 13.9    & 11.8   & 5.2   \\
\textit{+ unseen-mono} & \green{0.9}       & \green{0.1}       & \green{25.4}       & \green{2.6}    & \green{0.9}    & \green{10.4}     & \green{4.4}    & \green{6.3}   & \green{6.3}    \\ \midrule
mFTI-150       & 0.8       & 0.3       & 27.0       & 3.4    & 2.3    & 10.8    & 1.8     & 14.8   & 8.7    \\
\textit{+ unseen-mono} & \green{0.7}       & \green{0.2}       & \green{27.4}       & \green{0.5}    & \green{1.5}    & \green{12.0}    & \green{1.2}     & \green{5.1}   & \green{9.3}    \\ \midrule
\methodend-all               & 0.7       & 0.2       & 28.0       & 0.1    & 1.5    & 12.7    & 1.0     & 8.7   & 9.6   \\ \bottomrule
\end{tabular}
\caption{BLEU score, off-target ratio and oscillatory hallucination ratio before and after adding monolingual sentences to the finetuning corpus. Scores where adding monolingual sentences leads to improved quality are with \green{green} background. }
\label{tab:mono}
\end{table*}

\subsubsection{Adding Irrelevant Language Pairs Reduces SC, OT and OU Ratios} 

In Section \ref{sec:partial_scaling}, we show that additional language pairs in \method lead to improved BLEU scores even for the Unseen Both Sides partition. We provide an in-depth analysis here from the aforementioned fine-grained views. We plot the trends of translation and instruction-following performance, and the ratios of 4 specific instruction-following errors as the number of additional language pairs grows. The results are in Figure~\ref{fig:add_language_pairs}.

\paragraph{More Language Pairs Reduce Instruction-Following Errors and Improve Translation Performance.}
Firstly, we can see that as more language pairs are added to the training corpus, instruction-following errors on Unseen-both language pairs are gradually reduced, leading to improvements in BLEU scores. 
Comparing different language pairs, we can see that high- and medium-resource language pairs generally perform better than low-resource language pairs on all four types of errors. Since all these language directions are unseen when instruction finetuning, it highlights the importance of language skills acquired during the pretraining phase.

\paragraph{SC: Solved.} It can be observed that after adding about 30-60 language pairs, the model learns to avoid the SC problem, indicating this is a relatively easy problem to solve.

\paragraph{OU: Decreased to the level of \methodend-all. } We can further see that adding more language pairs is also effective for reducing OU errors, as the error ratios significantly decrease as the number of language pairs grows. Notably, after scaling the number of language pairs to 150, the OU ratios of three unseen language pairs are comparable to supervised full finetuning. This demonstrates the effectiveness of \methodend.

\paragraph{OT: Decreased, but not to a satisfactory level.} Turning to the OT ratio, we observe that it also decreases as the number of language pairs grows. However, even after scaling the number of language pairs to 150, the OT ratio still cannot be decreased to the level of \methodend-all.

\paragraph{OH: No effect.} Finally, we can see that with the increment in the number of language pairs, the OH ratio does not show a clear decreasing trend, which we will further discuss in the next section.

\subsubsection{Joint Training with Monolingual Generation Instructions Helps Reduce OH and OT Problems More Efficiently}
\label{sec:mono}
In the previous section, we find that the off-target (OT) and oscillatory hallucination (OH) on some language pairs cannot be fully solved to the level of \methodend-all by adding more irrelevant language pairs. We note that both problems are only related to the target language: the OT problem can be attributed to models' inability to relate target language names to the corresponding scripts of the language, and the OH problem might be caused by the poor modeling of the target languages. We hypothesize that finetuning models on 
 instructions of monolingual generation, i.e. given a language name, generate fluent sentences from that language, should help ease these problems.

To this end, we organize the monolingual sentences of the held-out languages into monolingual generation instructions. The template we adopt is ``$[l_i]: \mathbf{y}$''. We then finetune XGLM on the dataset compromised of translation instructions and these monolingual generation instructions. 

We report the BLEU score, OT ratio and OH ratio in Table~\ref{tab:mono}.  Firstly we can see that adding monolingual generation instructions for the three Unseen Both Side language pairs can help mitigate the OT and OH problem in most scenarios, leading to better translation performance. Notably, by combining more irrelevant language pairs and monolingual sentences, the gap between \methodend-150 with monolingual sentences and \methodend-all has significantly diminished, despite that the model has never seen parallel sentences of the tested language before.

\subsection{\method Improves Language Alignment via Pivot Languages}
\label{sec:language_alignment}

Besides the understanding of translation instruction, another crucial knowledge that models must grasp to carry out the instruction is the alignment between source and target languages. However, in scenarios where direct parallel sentences are not available, models have limited access to alignment information. This situation resembles the zero-shot setting commonly studied in multilingual translation research~\citep{gu-etal-2019-improved,zhang-etal-2020-improving,arivazhagan2019missing,liu-etal-2021-improving}. In this section, we aim to investigate the ability of \method to establish meaningful alignments through pivot languages in this scenario.

Specifically, for the three Unseen Both Sides language pairs X$\to$Y studied in the previous section, i.e. Ru$\to$Fr, Bg$\to$Ar and Ca$\to$Ta,
we start from the \methodend-150 setting, and add parallel sentences of X$\to$En and En$\to$Y to the training corpus. We then perform \method using these augmented corpora and evaluate the model's performance on test sentences that do not contain instruction-following errors. As knowledge of language alignments is the last requirement for carrying out translation instructions once the model has learned to execute translation instructions correctly, the performance on these sentences serves as a reliable indicator of the model's proficiency in language alignment.

The result is in Table~\ref{tab:pivot}. First, we can see that \methodend-150 and 8-shot ICL perform comparably, both significantly worse than \methodend-all. Since the tested three language pairs are unseen in \methodend-150, this indicates that similar to \methodend-150, the main role of ICL is to enhance the model's understanding of the translation behavior instead of source-target alignment knowledge. 
 
 However, after adding pivot parallel sentences, the model's performance (\textit{+pivot}) is significantly boosted. This demonstrates the potential of \method to leverage pivot languages to boost direct alignment between languages and improve translation performances.%

\begin{table}[t]
\centering
\begin{tabular}{rccc}
\toprule
                    & \textbf{Ru$\to$Fr} & \textbf{Bg$\to$Ar} & \textbf{Ca$\to$Ta} \\ \midrule
8-shot ICL          & 27.0           & 11.6            & 9.7                       \\
\methodend-all             & \textbf{28.2}           & \textbf{13.2}            & \underline{10.6}                       \\ \midrule
\methodend-150   & 27.5           & 11.6            & 9.2                       \\
\textit{+ pivot} & \underline{27.9}         & \underline{13.0}            & \textbf{10.8}                       \\ \bottomrule
\end{tabular}
\caption{Translation performance on test sentences without instruction-following errors. Best performances are in bold. The second-best performances are underlined.}
\label{tab:pivot}
\end{table}

\section{Related Works}

\subsection{LLMs for MT}
Machine translation researchers have widely recognized the potential of utilizing LLMs for MT, as these models acquire advanced language understanding skills during pretraining. The prevailing paradigm for leveraging LLMs for MT is in-context learning (ICL). For instance, \citet{lin-etal-2022-shot} demonstrated that providing 32 examples during translation can outperform GPT-3 and a supervised multilingual translation model. Other studies such as \citet{vilar-etal-2023-prompting}, \citet{agrawal-etal-2023-context}, and \citet{zhu2023multilingual} have investigated different factors that affect ICL's performance, including example quality, example selection strategy, and template sensitivity. Moreover, works such as \citet{hendy2023good} and \citet{jiao2023chatgpt} have studied the translation quality of various GPT-3 models and found their performances to be comparable to commercial translation systems on high-resource language pairs. In contrast to these works, our research focuses on exploring existing LLMs' translation ability by directly tuning them to follow translation instructions.

The most similar work to ours is \citet{jiao2023parrot}, which finetunes an open-source LLM LLaMA~\citep{touvron2023llama} on the mixes translation data and the \textit{alpaca} instruction dataset~\citep{alpaca} to make it a better translator. However, they mainly focus on the bilingual translation setting while our work investigates the multilingual generalization when finetuning LLMs to carry out translation instructions. 

\subsection{Generalization On Unseen Language Pairs}
Our work also has a close relation to zero-shot translation in the multilingual translation setting, where there are no direct parallel sentences between the source and target language. There are two major problems for zero-shot translation: generating correct languages and learning universal language representations. 

For the first problem, \citet{gu-etal-2019-improved,zhang-etal-2020-improving} leverage back-translation to add more target-language-related training data. \citet{arivazhagan2019missing,liu-etal-2021-improving} impose regularization on the encoder/decoder to make the model more aware of the target language. Unlike their works, we discuss the off-target problem in the context of LLMs, and find adding both irrelevant language pairs and additional monolingual sentences can ease the problem to a great extent.

For the second problem, previous works focus on learning language-agnostic representations through additional regularization of model representations~\citep{arivazhagan2019missing,pan-etal-2021-multilingual}, and consistency between semantic equivalent sentences~\citep{al-shedivat-parikh-2019-consistency,yang-etal-2021-multilingual}. Instead, our works mainly aim to reveal the helpfulness of multilingual finetuning LLMs for unseen language pairs by internalizing the pivot language information.

Furthermore, our discussion encompasses a more stringent version of zero-shot translation, where neither source nor target language sentences are present in the finetuning corpus. This demands a stronger generalization ability, as the model must effectively utilize the language knowledge acquired during pretraining and the translation task knowledge acquired during finetuning to generate high-quality translations.

\subsection{Instruction Finetuning}

Our work focuses on finetuning LLMs with instructions to improve zero-shot translation performance. Prior works have demonstrated that LLMs face great difficulty in achieving good performance in zero-shot settings when lacking few-shot examples. Nevertheless,  finetuning LLMs on a variety
of tasks can significantly improve zero-shot performance on several tasks. For instance, \citet{wei2022finetuned} aims to improve generalization in unseen tasks by performing instruction tuning. \citet{muennighoff2023crosslingual} further extend to finetune LLM by multilingual data instead of English data and find that multilingual finetuning leads to better performance on unseen tasks and unseen languages. \citet{chung2022scaling} explore instruction tuning from the perspective of the number of tasks in finetuning corpus and LLM size. \citet{chung2022scaling} found that scaling these factors can dramatically improve zero-shot performance.

In our work, we primarily focus on the translation performance of LLMs. We adopt a comprehensive approach to consider the factors mentioned above, including the scale of the finetuning corpus, the size of model parameters, and the language selection within the fine-tuning corpus, for a comprehensive analysis of the translation performance of the LLMs. Additionally, we conduct a detailed analysis of the model's understanding and execution capabilities in translation tasks after instruction finetuning.

\section{Conclusion}
In this paper, we explore Multilingual Finetuning with Translation Instructions (\methodend), to better unleash the translation ability of multilingual LLMs. Through extensive experiments, we demonstrate that by training on a mixture of 1000 sentences per language pair, \method achieves better performance than 8-shot ICL, indicating the untapped potential of translation ability in LLMs by previous works. 

Moreover, we systematically discuss the working mechanism of \method by analyzing it from the view of instruction-following. Our experiments demonstrate that \method helps the model better follow the instruction by introducing more language pairs and monolingual sentences, and enhances the direct language alignment by learning from pivot language pairs. 

Our paper also unveils remaining translation issues when adopting LLMs for zero-shot machine translation, i.e. over/under translation, oscillatory hallucination, and mistranslation caused by incorrect alignments. Future works should focus on acquiring more language knowledge from the pretraining phase and designing better regularization terms to solve these problems.

\section*{Acknowledgement}
We would like to thank the anonymous reviewers and the editor for their insightful comments. Shujian Huang is the corresponding author. This work is supported by National Science Foundation of China (No. 62376116, 62176120), the Liaoning Provincial Research Foundation for Basic Research (No. 2022-KF-26-02).

\bibliography{anthology,tacl2021}
\bibliographystyle{acl_natbib}


\appendix

\appendix

\begin{figure}[t!]
\centering
\includegraphics[width=0.95\linewidth]{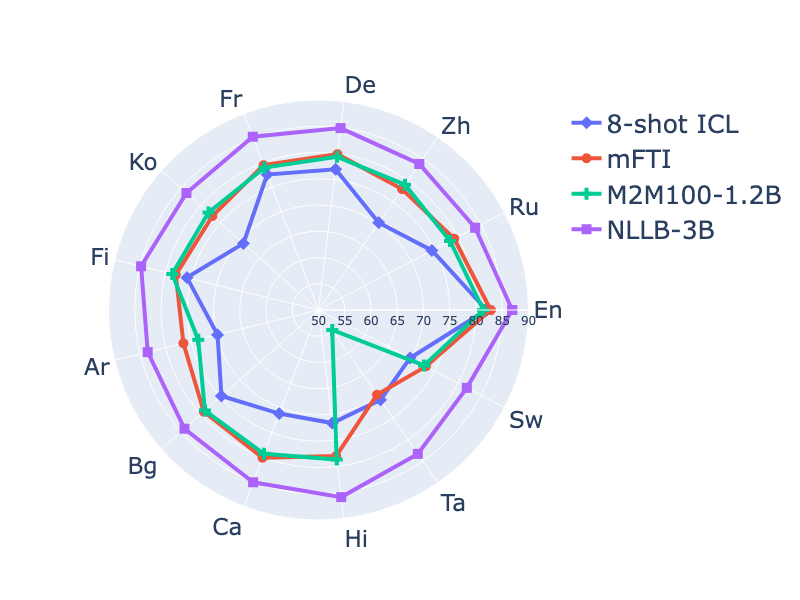}
\caption{Comparison of \method with conventional supervised machine translation models. Performances are evaluated in COMET. }
\label{fig:supervised_comet}
\end{figure}

\begin{table*}[t]
\centering
\small
\begin{tabular}{lcl|c|c} 

\toprule
&  & & BLOOM-7B and LLaMA-7B & XGLM-7.5B \\ \midrule
Method      & \multicolumn{2}{l|}{Hyperparameter} & Value   & Value \\ \midrule
\multirow{8}{*}{LoRA}  & \multirow{4}{*}{LoRA} & modules   & query, key, value   & query, key, value  \\
                       & & rank                            & 4                           & 4 \\
                       & & scaling factor                  & 32                          & 32 \\
                       & & dropout                         & 0.1                         & 0.1 \\
                       &  \multicolumn{2}{l|}{learning rate} & 5e-5                        & 5e-4 \\
                       &  \multicolumn{2}{l|}{batch size}    & 80                          & 80 \\
                       &  \multicolumn{2}{l|}{training steps} & 5000                       & 5000 \\
                       &  \multicolumn{2}{l|}{evaluation frequency}& 500                   & 500 \\  \midrule   
\multirow{4}{*}{Full-finetuning}  &  \multicolumn{2}{l|}{learning rate}    & 1e-5          & 5e-6 \\
                                  &  \multicolumn{2}{l|}{batch size}    & 80               & 80 \\
                                  &  \multicolumn{2}{l|}{training steps} & 2500            & 2000 \\
                                  &  \multicolumn{2}{l|}{evaluation frequency}& 250        & 250 \\ \bottomrule
\end{tabular}
\caption{Hyperparameter configurations of LoRA
and Full-fine-tuning in LLMs}
\label{tab:hyper}
\end{table*}

\section*{Appendix}
\label{sec:appendix}

\section{Full results}
\label{appendix: full results}
Table \ref{tab:full_res} shows all 156 language pair results of 8-shot ICL and \method on XGLM, evaluated by both BLEU and COMET.

\section{Training and Evaluation details of mfti-156 in different LLMs}
\label{appendix:hyperparameters}

The distribution of pretraining corpus varies across different LLMs,  hence we adopt diverse hyperparameters in Table \ref{tab:hyper}. Moreover, we conduct evaluations on LLMs at regular steps (250 for Full-finetuning and 500 for Lora) during the training phase, selecting the best-performing result in the end.

\section{Comparison of \method and supervised machine translation models evaluated by COMET}
\label{sec:supervised_comet}
We present the comparison of \method and supervised MT models evaluated by COMET in Figure~\ref{fig:supervised_comet}. It can be seen that when evaluated by COMET, \methodend's performance is comparable to M2M-1.2B, yet still substantially underperforms NLLB-3B.

\section{Additional language pairs for \method}
\label{appendix:Additional language}

We construct the additional language pairs in Section~\ref{sec:partial_scaling} from the other 17 languages covered in the pretraining corpus of XGLM, including Spanish, Greek, Portuguese, Japanese, Vietnamese, Urdu, Thai, Turkish, Telugu, Italian, Haitian, Creole, Basque, Indonesian, Estonian and Bangali.

\begin{table*}[!b]
\centering
\tiny
\begin{tabular}{c|clccccccccccccccc}
\toprule
  &   &  & en & de & fr & ca & fi & ru & bg & zh & ko & ar & sw & hi & ta  & avg   \\ \midrule
\multirow{4}{*}{en}  &\multirow{2}{*}{bleu}&  
                     8-shot ICL & - & 29.4 & 38.5 & 37.2 & 25.1 & 25.6 & 34.8 & \textbf{17.5} & 15.6 & 15.7 & \textbf{22.2} & 19.9 & \textbf{10.1} & 24.3 \\ 
                     & &\method & - & \textbf{30.4} & \textbf{39.7} & \textbf{37.5} & \textbf{25.5} & \textbf{26.3} & \textbf{35.2} & 16.6 & \textbf{16.2} & \textbf{16.6} & 22.1 & \textbf{21.7} & 9.1 & \textbf{24.7}  \\ 
                     \cmidrule(lr){2-17}
                     & \multirow{2}{*}{comet}&  8-shot ICL & - &82.7 &83.2 &84.5 &89.8 &85.4 &87.7 &\textbf{81.4} &\textbf{82.7} &80.2 &79.1 &70.6 &\textbf{77.1} &82.0  \\ 
                     & & \method & - &\textbf{84.3} &\textbf{85.1} &\textbf{85.2} &\textbf{90.8} &\textbf{85.8} &\textbf{88.4} &81.1 &81.8 &\textbf{81.8} &\textbf{80.8} &\textbf{73.0} &75.2 &\textbf{82.8} \\ \midrule

\multirow{4}{*}{de}  &\multirow{2}{*}{bleu} 
                     &  8-shot ICL & 38.0 & - & 28.8 & 17.6 & \textbf{20.9} & 19.7 & 25.3 & 12.4 & 7.7 & 8.9 & \textbf{14.8} & \textbf{16.4} & \textbf{8.9} & 18.3   \\ 
                     &&  \method & \textbf{41.3} & - & \textbf{31.0} & \textbf{27.7} & 20.2 & \textbf{21.0} & \textbf{26.6} & \textbf{12.9} & \textbf{14.3} & \textbf{16.9} & 11.7 & 16.0 & 5.2 & \textbf{20.4}  \\ 

                     \cmidrule(lr){2-17}
                     
                     & \multirow{2}{*}{comet}&  8-shot ICL & 86.3 &- &80.0 &77.7 &86.7 &82.6 &83.9 &77.8 &69.4 &69.3 &73.7 &65.0 &\textbf{71.6} &77.0    \\  
                     
                     & & \method & \textbf{90.1} &- &\textbf{85.3} &\textbf{82.4} &\textbf{88.5} &\textbf{87.2} &\textbf{86.1} &\textbf{78.4} &\textbf{77.8} &\textbf{78.6} &\textbf{74.7} &\textbf{66.1} &63.7 &\textbf{79.9}\\ \midrule
                     
\multirow{4}{*}{fr}  &\multirow{2}{*}{bleu} 
                     & few-shot & 39.3 & 15.8 & - & 29.0 & \textbf{19.5} & 20.5 & 27.2 & 12.3 & 6.5 & 10.7 & \textbf{18.1} & \textbf{14.3} & \textbf{8.9} & 18.5   \\ 
                     & & \method & \textbf{41.7} & \textbf{23.8} & - & \textbf{33.9} & 18.6 & \textbf{22.0} & \textbf{27.5} & \textbf{16.7} & \textbf{11.6} & \textbf{13.7} & 15.0 & 14.0 & 6.6 & \textbf{20.4} \\ 
                     
                     \cmidrule(lr){2-17}

                     & \multirow{2}{*}{comet}&  8-shot ICL & 86.1 &75.4 &- &81.7 &\textbf{86.5} &83.4 &86.3 &77.0 &68.3 &73.3 &\textbf{76.6} &63.5 &\textbf{73.5} &77.6   \\  
                     & & \method & \textbf{86.7} &\textbf{84.8} &- &\textbf{86.2} &85.5 &\textbf{85.1} &\textbf{88.6} &\textbf{78.1} &\textbf{77.7} &\textbf{78.7} &74.7 &\textbf{64.1} &63.9 &\textbf{79.5} \\ \midrule

\multirow{4}{*}{ca}  &\multirow{2}{*}{bleu} 
                     &  8-shot ICL & 41.0 & 19.5 & 33.5 & - & 9.2 & 9.0 & 23.5 & 11.1 & 4.2 & 1.8 & 13.7 & 10.8 & 8.4 & 15.5   \\ 
                     &  & \method & \textbf{42.6} & \textbf{22.4} & \textbf{35.6} & - & \textbf{18.2} & \textbf{21.4} & \textbf{27.2} & \textbf{12.6} & \textbf{11.4} & \textbf{13.5} & \textbf{16.3} & \textbf{14.2} & \textbf{9.6} & \textbf{20.4}  \\ 

                     \cmidrule(lr){2-17}
                     & \multirow{2}{*}{comet}&  
                     8-shot ICL & 86.4 &76.3 &82.2 &- &70.1 &61.6 &81.4 &77.5 &60.5 &57.2 &69.8 &59.2 &\textbf{71.4} &71.1   \\  
                     & & \method & \textbf{87.5} &\textbf{81.4} &\textbf{85.7} &- &\textbf{88.7} &\textbf{85.3} &\textbf{86.3} &\textbf{79.7} &\textbf{77.3} &\textbf{78.6} &\textbf{77.6} &\textbf{66.0} &66.9 &\textbf{80.1}  \\ \midrule

\multirow{4}{*}{fi}  &\multirow{2}{*}{bleu} 
                     &  8-shot ICL & 29.0 & \textbf{17.9} & 23.9 & 18.4 & - & 15.9 & 12.7 & \textbf{12.2} & 10.0 & 8.9 & 9.2 & 10.3 & 3.9 & 14.4  \\ 
                     &&  \method & \textbf{30.6} & 17.6 & \textbf{24.4} & \textbf{23.2} & - & \textbf{17.4} & \textbf{20.7} & 12.0 & \textbf{11.7} & \textbf{9.7} & \textbf{14.9} & \textbf{12.9} & \textbf{4.1} & \textbf{16.6}  \\ 
                     
                     \cmidrule(lr){2-17}
                     & \multirow{2}{*}{comet}&  
                     8-shot ICL & 87.1 &\textbf{80.8} &81.2 &80.4 &- &82.2 &74.7 &\textbf{79.1} &76.6 &74.8 &70.9 &60.5 &61.7 &75.8    \\  
                     & & \method & \textbf{88.0} &79.4 &\textbf{82.0} &\textbf{81.7} &- &\textbf{85.2} &\textbf{84.3} &78.3 &\textbf{76.7} &\textbf{76.0} &\textbf{73.4} &\textbf{66.6} &\textbf{64.8} &\textbf{78.0}   \\ \midrule
                     
\multirow{4}{*}{ru}  &\multirow{2}{*}{bleu} 
                     &  8-shot ICL & 30.9 & 19.3 & 25.7 & 15.6 & 4.2 & - & 25.5 & 12.0 & 7.7 & 7.9 & 7.8 & 12.2 & \textbf{6.8} & 14.6  \\ 
                     &&  \method & \textbf{32.5} & \textbf{20.4} & \textbf{28.0} & \textbf{25.7} & \textbf{17.6} & - & \textbf{29.5} & \textbf{12.4} & \textbf{10.1} & \textbf{14.9} & \textbf{12.2} & \textbf{13.2} & 6.2 & \textbf{18.6}  \\ 
                     
                     \cmidrule(lr){2-17}
                     
                     & \multirow{2}{*}{comet}&  
                     8-shot ICL & 83.7 &78.5 &79.1 &76.9 &72.6 &- &88.4 &\textbf{77.1} &72.1 &70.4 &64.8 &62.0 &\textbf{66.7} &74.4   \\  
                     & & \method & \textbf{86.2} &\textbf{80.6} &\textbf{84.1} &\textbf{83.0} &\textbf{88.1} &- &\textbf{91.7} &76.3 &\textbf{76.4} &\textbf{78.4} &\textbf{75.6} &\textbf{67.0} &63.4 &\textbf{79.2}   \\ \midrule
                     
\multirow{4}{*}{bg}  &\multirow{2}{*}{bleu} 
                     &  8-shot ICL & 35.7 & 20.4 & 26.6 & 14.5 & 14.5 & 22.8 & - & 12.5 & 8.9 & 10.7 & 10.0 & 6.8 & 1.6 & 15.4   \\ 
                     &&  \method & \textbf{37.6} & \textbf{21.2} & \textbf{30.2} & \textbf{28.1} & \textbf{17.1} & \textbf{24.7} & - & 12.5 & \textbf{10.2} & \textbf{12.7} & \textbf{12.9} & \textbf{14.4} & \textbf{4.6} & \textbf{19.2} \\ 

                     \cmidrule(lr){2-17}
                     
                     & \multirow{2}{*}{comet}&  
                     8-shot ICL & 85.8 &78.7 &79.8 &78.2 &81.0 &86.8 &- &77.1 &74.4 &74.8 &69.5 &53.6 &57.1 &74.7   \\  
                     & & \method &\textbf{88.0} &\textbf{83.4} &\textbf{83.0} &\textbf{84.8} &\textbf{87.3} &\textbf{88.9} &- &\textbf{77.4} &\textbf{74.5} &\textbf{78.4} &\textbf{74.8} &\textbf{66.6} &\textbf{63.5} &\textbf{79.2}   \\ \midrule                    

\multirow{4}{*}{zh}  &\multirow{2}{*}{bleu} 
                     &  8-shot ICL & 21.9 & 6.5 & 10.4 & 6.4 & 3.6 & 1.9 & 14.7 & - & 9.6 & 3.0 & 7.6 & \textbf{13.3} & \textbf{9.6} & 9.0   \\ 
                     &&  \method & \textbf{23.4} & \textbf{11.5} & \textbf{18.7} & \textbf{25.2} & \textbf{12.3} & \textbf{12.8} & \textbf{15.4} & - & \textbf{12.4} & \textbf{8.5} & \textbf{9.5} & 12.9 & 4.6 & \textbf{13.9}  \\ 

                     \cmidrule(lr){2-17}
                     
                     & \multirow{2}{*}{comet}&  
                     8-shot ICL & 82.5 &66.4 &70.9 &70.1 &65.4 &56.7 &81.0 &- &77.7 &62.1 &69.3 &64.6 &\textbf{75.6} &70.2 &  \\  
                     & & \method & \textbf{86.1} &\textbf{76.4} &\textbf{78.9} &\textbf{80.4} &\textbf{84.7} &\textbf{83.7} &\textbf{84.3} &- &\textbf{79.9} &\textbf{76.2} &\textbf{73.0} &\textbf{67.3} &64.8 &\textbf{78.0}  \\ \midrule   

\multirow{4}{*}{ko}  &\multirow{2}{*}{bleu}
                     &  8-shot ICL & 21.1 & \textbf{11.2} & 15.0 & 7.3 & 2.8 & 3.1 & 9.9 & 11.9 & - & 3.0 & 2.4 & 0.7 & 0.4 & 7.4   \\ 
                     &&  \method & \textbf{22.9} & 10.3 & \textbf{16.0} & \textbf{15.5} & \textbf{10.8} & \textbf{10.4} & \textbf{13.4} & \textbf{12.1} & - & \textbf{9.8} & \textbf{9.1} & \textbf{13.2} & \textbf{6.3} & \textbf{12.5}  \\ 

                      \cmidrule(lr){2-17}
                     
                     & \multirow{2}{*}{comet}&  
                     8-shot ICL    & 83.6 &74.3 &\textbf{75.5} &73.0 &66.7 &58.4 &74.5 &78.8 &- &62.4 &58.8 &56.4 &66.6 &69.1 &\\  
                     & & \method & \textbf{86.2} &\textbf{75.9} &74.8 &\textbf{78.6} &\textbf{84.6} &\textbf{82.1} &\textbf{82.1} &\textbf{79.4} &- &\textbf{72.9} &\textbf{73.8} &\textbf{64.3} &\textbf{69.8} &\textbf{77.0}  \\ \midrule  

\multirow{4}{*}{ar}  &\multirow{2}{*}{bleu}
                     &  8-shot ICL & 28.2 & 12.7 & 21.3 & 8.8 & 4.3 & 10.0 & 19.0 & \textbf{9.4} & 1.7 & - & 4.3 & 10.4 & \textbf{7.2} & 11.4   \\ 
                     &&  \method & \textbf{30.6} & \textbf{13.7} & \textbf{23.2} & \textbf{24.8} & \textbf{14.0} & \textbf{14.9} & \textbf{21.9} & 9.3 & \textbf{8.3} & - & \textbf{11.9} & \textbf{11.4} & 4.1 & \textbf{15.7}  \\ 

                     \cmidrule(lr){2-17}
                     
                     & \multirow{2}{*}{comet}&  
                     8-shot ICL    &82.5 &72.5 &76.0 &73.4 &65.8 &69.6 &80.0 &\textbf{74.4} &56.2 &- &61.0 &58.1 &\textbf{67.9} &69.8  \\  
                     & & \method & \textbf{86.6} &\textbf{75.4} &\textbf{79.4} &\textbf{82.9} &\textbf{84.0} &\textbf{83.9} &\textbf{84.5} &73.1 &\textbf{71.6} &- &\textbf{72.4} &\textbf{61.6} &62.5 &\textbf{76.5}   \\ \midrule  
                     
\multirow{4}{*}{sw}  &\multirow{2}{*}{bleu}
                     &  8-shot ICL & \textbf{32.2} & 14.2 & \textbf{23.2} & 17.5 & \textbf{10.5} & \textbf{13.2} & 12.4 & \textbf{9.5} & 7.1 & 9.7 & - & 8.5 & 2.7 & 13.4   \\ 
                     &&  \method & 32.1 & \textbf{14.9} & 21.0 & \textbf{20.4} & 10.4 & 12.5 & \textbf{16.3} & 8.6 & \textbf{9.0} & 10.4 & - & \textbf{12.7} & \textbf{4.8} & \textbf{14.4}  \\ 
                     
                     \cmidrule(lr){2-17}
                     
                     & \multirow{2}{*}{comet}&  
                     8-shot ICL & 80.9 &\textbf{70.9} &\textbf{73.2} &73.6 &73.4 &\textbf{76.6} &68.0 &\textbf{72.2} &70.2 &71.3 &- &54.2 &51.6 &69.7      \\  
                     & & \method & \textbf{82.7} &70.0 &72.6 &\textbf{75.8} &\textbf{76.6} &76.2 &\textbf{81.0} &70.0 &\textbf{72.5} &\textbf{74.6} &- &\textbf{62.1} &\textbf{63.0} &\textbf{73.1}   \\ \midrule 
                     
\multirow{4}{*}{hi}  &\multirow{2}{*}{bleu}
                     &  8-shot ICL & 23.7 & 12.8 & 15.6 & 8.4 & 9.5 & 11.3 & 9.9 & \textbf{11.3} & 11.0 & 6.0 & 5.0 & - & 0.2 & 10.4   \\ 
                     &&  \method & \textbf{28.0} & 12.8 & \textbf{17.8} & \textbf{16.9} & \textbf{12.1} & \textbf{12.0} & \textbf{16.3} & 10.8 & \textbf{12.2} & \textbf{8.9} & \textbf{10.9} & - & \textbf{10.1} & \textbf{14.1}  \\

                     \cmidrule(lr){2-17}
                     
                     & \multirow{2}{*}{comet}&  
                     8-shot ICL & 82.5 &74.6 &74.9 &74.1 &78.1 &78.5 &66.4 &\textbf{76.9} &76.7 &68.7 &62.8 &- &45.7 &71.7 \\  
                     & & \method &\textbf{86.1} &\textbf{76.2} &\textbf{75.3} &\textbf{79.9} &\textbf{82.2} &\textbf{81.3} &\textbf{81.1} &75.9 &\textbf{77.9} &\textbf{75.0} &\textbf{73.8} &- &\textbf{71.7} &\textbf{78.0}  \\ \midrule 
                     
\multirow{4}{*}{ta}  &\multirow{2}{*}{bleu}
                     &  8-shot ICL & \textbf{16.1} & \textbf{8.8} & \textbf{11.2} & 5.7 & \textbf{6.9} & \textbf{9.0} & \textbf{8.9} & \textbf{8.3} & \textbf{8.7} & \textbf{5.5} & 3.1 & 6.6 & - & 8.2   \\ 
                     &&  \method & 16.0 & 7.7 & 9.5 & \textbf{10.4} & 5.9 & 6.3 & 8.1 & 6.6 & 7.7 & 5.4 & \textbf{6.4} & \textbf{13.3} & - & \textbf{8.6}  \\ 

                      \cmidrule(lr){2-17}
                     
                     & \multirow{2}{*}{comet}&  
                     8-shot ICL & 78.5 &\textbf{70.7} &\textbf{71.3} &\textbf{70.6} &74.2 &\textbf{77.0} &72.3 &\textbf{73.2} &\textbf{75.6} &\textbf{71.1} &61.4 &53.3 &- &\textbf{70.8}  \\  
                     & & \method & \textbf{78.6} &64.3 &66.0 &70.1 &\textbf{74.4} &71.6 &\textbf{72.9} &68.4 &69.0 &69.1 &\textbf{67.1} &\textbf{63.4} &- &69.6 \\ \midrule 

\multirow{4}{*}{avg} &\multirow{2}{*}{bleu}
                     &  8-shot ICL & 29.8 & 15.7 & 22.8 & 15.5 & 10.9 & 13.5 & 18.7 & 11.7 & 8.2 & 7.7 & 9.9 & 10.9 & 5.7 & 13.9  \\ 
                     &&  \method & \textbf{31.6} & \textbf{17.2} & \textbf{24.6} & \textbf{24.1} & \textbf{15.2} & \textbf{16.8} & \textbf{21.5} & \textbf{11.9} & \textbf{11.3} & \textbf{12.1} & \textbf{12.7} & \textbf{14.1} & \textbf{6.3} & \textbf{16.9}  \\

                     \cmidrule(lr){2-17}
                     
                     & \multirow{2}{*}{comet}&  
                     8-shot ICL & 83.8 &75.2 &77.3 &76.2 &75.9 &74.9 &78.7 &\textbf{76.9} &71.7 &69.6 &68.1 &60.1 &65.5 &73.4   \\  
                     & & \method &  \textbf{86.1} &\textbf{77.7} &\textbf{79.4} &\textbf{80.9} &\textbf{84.6} &\textbf{83.0} &\textbf{84.3} &76.3 &\textbf{76.1} &\textbf{76.5} &\textbf{74.3} &\textbf{65.7} &\textbf{66.1} &\textbf{77.7}  \\ \bottomrule

\end{tabular}
\caption{Translation performance of 8-shot ICL and \method based on XGLM-7.5B on FLORES-101~(test). Source language in rows, target language in columns.}
\label{tab:full_res}
\end{table*}


\end{document}